\documentclass[letterpaper, 10 pt, conference]{ieeeconf}  % Comment this line out
\IEEEoverridecommandlockouts                              % This command is only
\usepackage{xcolor}
\newcommand{\revised}[1]{\textcolor{black}{#1}}
\usepackage[utf8]{inputenc}   % 输入使用 UTF-8 编码
\usepackage[T1]{fontenc}      % 支持西欧语言的输出编码

\usepackage{amsmath,amsfonts}

\usepackage{bm}
\usepackage{algpseudocode}
\usepackage{algorithm}
\usepackage[caption=false]{subfig}
\usepackage{tikz}
\usetikzlibrary{arrows.meta, automata, positioning}

\usepackage{array}
\usepackage{textcomp}
\usepackage{stfloats}
\usepackage{url}
\usepackage{verbatim}
\usepackage{graphicx}

\usepackage{booktabs}
\usepackage{tabularx}

\usepackage{siunitx}

\usepackage{cite}
\makeatletter
\let\NAT@parse\undefined
\makeatother

\usepackage[colorlinks,linkcolor=blue,anchorcolor=black,citecolor=blue,urlcolor=blue,hyperfootnotes=true]{hyperref}   % hyperref must be put at the end!
\usepackage[all]{hypcap} 
\title{\LARGE \bf
Shared-Control Teleoperation of Omnidirectional Aerial Robots for Continuous Valve Rotation
}

\author{Jinjie Li, Yicheng Chen, Haokun Liu, and Moju Zhao% <-this % stops a space
\thanks{All members come from DRAGON Lab, Department of Mechanical Engineering, The University of Tokyo, Tokyo 113-8656, Japan
        (email: {\tt\small jinjie-li@dragon.t.u-tokyo.ac.jp})}%
}

\begin{document}
\bstctlcite{IEEEtran:BSTcontrol}  %%% MUST for the use of my_config.bib

\maketitle
\thispagestyle{empty}
\pagestyle{empty}

%%%%%%%%%%%%%%%%%%%%%%%%%%%%%%%%%%%%%%%%%%%%%%%%%%%%%%%%%%%%%%%%%%%%%%%%%%%%%%%%
\begin{abstract}

Omnidirectional aerial robots offer full 6-DoF independent control over position and orientation, making them popular for aerial manipulation. Although there have been advancements in robotic autonomy, human operation remains essential in complex aerial environments. Previously, we proposed an aerial teleoperation system \cite{liSixDoFHandBasedTeleoperation2025} that brings the rotational flexibility of human hands into the unbounded aerial workspace, which was evaluated through a vertically mounted valve-turning task.
% Additionally, the dexterity of human fingers should be exploited for more engaged interaction.
% The position and orientation of omnidirectional aerial robots can be independently controlled in any directions, making them increasingly popular for aerial manipulation. Considering the complex environment faced in the air, teleoperation for these robots is still required. 
% However, the existed teleoperation approaches for multi-rotors cannot account for the additional DoFs brought by rotation. Furthermore, the freedom of fingers are always ignored in previous work.
However, due to the limited range of human forearm’s motion, the robot could not continuously rotate the valve, hindering its application in industry. In this work, we propose an approach to initiate continuous rotation when the forearm angle exceeds a threshold. This switch of control authority exemplifies the concept \textit{shared control} from human-robot interaction, marking a further step in aerial teleoperation.

% In this work, we propose an aerial teleoperation system that brings the \revised{rotational flexibility} of human hands into the unbounded aerial workspace. Our system includes two motion-tracking marker sets---one on the shoulder and one on the hand---along with a data glove to capture hand gestures. Using these inputs, we design four interaction modes for different tasks, including \textit{Spherical Mode} and \textit{Cartesian Mode} for long-range moving, \textit{Operation Mode} for precise manipulation, as well as \textit{Locking Mode} for temporary pauses, where the hand gestures are utilized for seamless mode switching.
% We evaluate our system on a \revised{vertically mounted} valve-turning task in \revised{the} real world, demonstrating how each mode contributes to effective aerial manipulation. 
% % Results show that: (1) \textit{Spherical Mode} intuitively extends the human hand for large-range movements, while \textit{Cartesian Mode} facilitates precise navigation along structured paths. (2) For fine manipulation, \textit{Operation Mode} directly maps human hand motions to the robot, while \textit{Locking Mode} is crucial when the operator’s line of sight is obstructed. 
% This interaction framework bridges human dexterity with aerial robotics, paving the way for enhanced aerial teleoperation in unstructured environments.

\end{abstract}

%%%%%%%%%%%%%%%%%%%%%%%%%%%%%%%%%%%%%%%%%%%%%%%%%%%%%%%%%%%%%%%%%%%%%%%%%%%%%%%%
\section{Introduction}

% For millions of years, the manipulation space of we human's hands are limited by the length of our arms. 

% We humans continually aspire for robots to perform more powerful and precise operations over larger areas. Although traditional robot manipulators are designed to deliver greater strength and sometimes higher precision, their working spaces remain limited by the reach of their arms—much like human arms. To overcome these constraints, 

Aerial manipulation is gaining popularity due to its ability to operate freely in 3D space, particularly in high-altitude environments that are difficult for humans to access \cite{ollero_past_2022, zhao_forceful_2022, nishio_design_2024}.
Although fully autonomous manipulation remains the ultimate goal for robotics researchers, the complexity of the aerial environment necessitates keeping a human operator in the loop \cite{darvish_teleoperation_2023}. In our previous work \cite{liSixDoFHandBasedTeleoperation2025}, we proposed a six-degree-of-freedom teleoperation framework for aerial manipulation. Specifically, we designed four interaction modes for different tasks, including \textit{Spherical Mode} and \textit{Cartesian Mode} for long-range movement, \textit{Operation Mode} for precise manipulation, and \textit{Locking Mode} for temporary pauses,where hand gestures were utilized for seamless mode switching. The proposed framework was evaluated through a valve-turning task\footnote{The video of previous work: \href{https://youtu.be/n0IQEnjPzrw}{https://youtu.be/n0IQEnjPzrw}}, and to the best of our knowledge, this was the first time the rotational dexterity of the human wrist was intuitively demonstrated on an omnidirectional aerial robot. However, due to the limitations of both the controller and the physical structure of human forearm, we were not able to achieve continuous rotation.

In this work, we introduce the concept of \textit{shared control} \cite{selvaggio_autonomy_2021} from the human–machine interaction (HMI) domain into aerial operations and design a mechanism to temporarily shift control authority to the robot for continuous valve turning. The approach is briefly introduced as follows, and its performance is evaluated in a real-world experiment.

\setlength{\textfloatsep}{8pt plus 1.0pt minus 2.0pt}
\begin{figure}[t]
    \centerline{\includegraphics[trim=0 0 0 0,clip,width=3.3in]{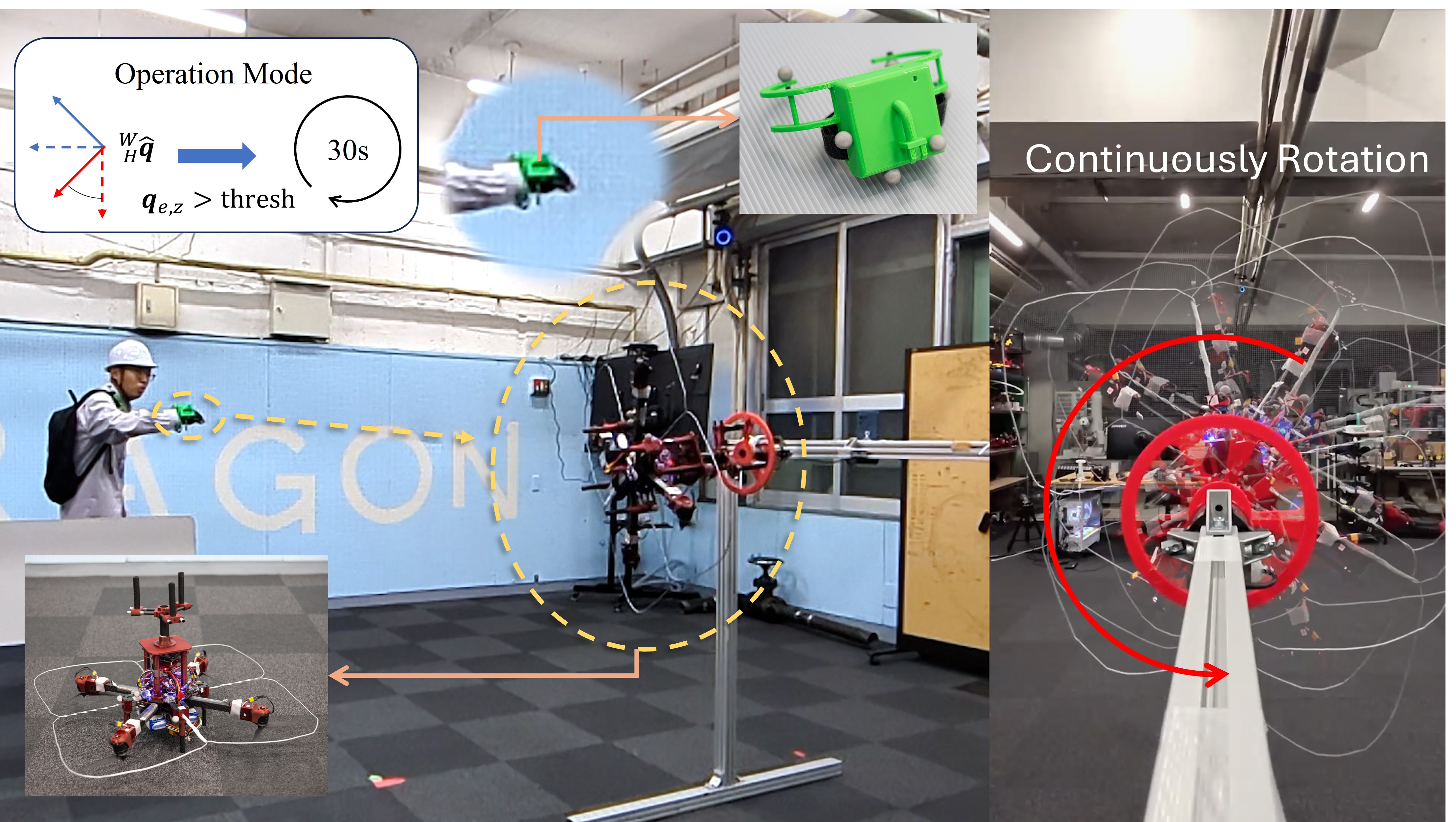}} 
    % 这个trim可以裁掉图片的白边
    % trim=left bottom right top
    \vspace*{-2mm}
    \caption{\textbf{A human operator controls an omnidirectional aerial robot to continuously turn a vertically mounted valve.} The robot is in \textit{Operation Mode}, where its target pose is in general mapped directly from the operator’s hand pose. When the forearm rotation exceeds a threshold, continuous 360° rotation is triggered.
    % The position during this process is still controlled by human hand for safety.
    % The current mode is visually fed back to the user through a display.
    The link of video:
    \href{https://youtu.be/JuPAP6E4yR4}{https://youtu.be/JuPAP6E4yR4}.
    }
    \label{fig:teleop_real}
\end{figure}

\section{Omnidirectional Aerial Teleoperation}
 \label{sec:system}

The teleoperation system for an omnidirectional aerial robot consists of four main elements, as shown in Fig.\;\ref{fig:teleop_real}: (1) \textit{a human operator}, (2) \textit{motion-tracking marker sets} \revised{to localize the hand}, (3) \textit{a data glove} \revised{to capture hand gestures}, and (4) \textit{an omnidirectional aerial robot}.

% \setlength{\textfloatsep}{8pt plus 1.0pt minus 2.0pt}
% \begin{figure}[t]
%     \centerline{\includegraphics[trim=5.5cm 2.5cm 9cm 2.5cm,clip,width=3.2in]{figs/workflow_v2.pdf}} 
%     % 这个trim可以裁掉图片的白边
%     % trim=left bottom right top
%     \vspace*{-1mm}
%     \caption{\textbf{Illustration of our teleoperation system.}
%     % A human operator wears two marker sets—one on the shoulder and one on the hand—along with a data glove on one hand. The poses of marker sets are tracked for the interaction framework to control the robot, and the glove data are used to recognize hand gestures for mode switching. The teleoperation scenario, including the robot and the operated object, is visually fed back to the operator. 
%     Since the limit of physiological structure in human forearm, we introduce a shared control mechanism for continuous rotation: if the rotation of forearm is beyond a threshold, the control of robot's orientation is shifted from human to program, resulting a continuous rotation for one round.
%     }
%     \label{fig:system}
% \end{figure}

To thoroughly exploit the omnidirectionality of the human hand, an aerial robot capable of hovering in any orientation is required. Here, we use an omnidirectional upgrade of the tiltable quadrotor presented in \cite{li_servo_2024}. The control method is actuator-level NMPC, with an upgrade to handle singularities during continuous rotation.

We define the body frame $\left\{\mathcal{B}\right\}$ of the drone at its center of gravity (CoG). With the flight controller, the robot can be simplified as a first-order model:
\begin{subequations} \label{eq:rb}
\vspace{-2mm}
\begin{align}
{\dot{\boldsymbol{p}}_B} &= 1/\boldsymbol{t}_{p} \cdot \left( {\boldsymbol{p}_B}_c - {\boldsymbol{p}_B} \right), \label{eq:rb1} \\[5pt]
{^W_B\dot{\boldsymbol{q}}} &= 1/\boldsymbol{t}_{q} \cdot \left( {^W_B\boldsymbol{q}^*} \circ {^W_B\boldsymbol{q}_c} \right), \label{eq:rb3}
\end{align}
\end{subequations}
where $\boldsymbol{p}_B$, ${\boldsymbol{p}_B}_c$, ${^W_B\boldsymbol{q}}$, and ${^W_B\boldsymbol{q}}_c$ represent current position, target position, current quaternion, and target quaternion, respectively, $\circ$ refers to quaternion multiplication, $^*$ refers to conjugation, as well as $\boldsymbol{t}_{p}$ and $\boldsymbol{t}_{q}$ denote time constants determined by the control performance. The estimated position and orientation are denoted as $\hat{\boldsymbol{p}}_B$ and ${^W_B\hat{\boldsymbol{q}}}$.

We also require the hand's position $\hat{\boldsymbol{p}}_H$ and orientation ${^W_H\hat{\boldsymbol{q}}}$ to determine the robot's target pose, which is obtained by Motion Capture System. The interaction framework \revised{functions as} a planning module, aiming to calculate ${\boldsymbol{p}_B}_c$ and ${^W_B\boldsymbol{q}}_c$ based on the operator's action.

% \( \left\{\mathcal{S}\right\} \) represents the shoulder's local coordinate system.
% The position of the origin of \( \left\{\mathcal{H}\right\} \) in the world frame \( \left\{\mathcal{W}\right\} \) is expressed as \( ^{W}\boldsymbol{p}_H \).

% \subsection{Data Glove}

% Data gloves can capture the movement of each finger, which is essential for leveraging the dexterity of the human hand. Generally, data gloves are equipped with multiple fabric-based sensors that measure the expansion and contraction of the material as the fingers move. In our system, the stretch values from all knuckles, denoted as $\boldsymbol{h}$, are used to determine the hand gesture.

\section{Hand-Based Interaction}  \label{sec:interaction}

% First, \textit{Operation Mode} and \textit{Locking Mode} are introduced for precise manipulation. Then, \textit{Spherical Mode} and \textit{Cartesian Mode} are introduced for long-range moving. Finally, a gesture-based mode switch method (depicted in Fig.\;\ref{fig:mode_switch_logic}) and the corresponding visual feedback are presented.
% In our previous work \cite{liSixDoFHandBasedTeleoperation2025}, we introduce \textit{Spherical Mode} and \textit{Cartesian Mode} for long-range moving, \textit{Operation Mode} for precise manipulation, as well as \textit{Locking Mode} to freeze the drone’s pose. Since the focus of this paper is the continuous rotation in \textit{Operation Mode}, we only introduce this mode and the triggering mechanism for rotation.

We define an \textit{Operation Mode} in \cite{liSixDoFHandBasedTeleoperation2025}, in which we aim to establish an intuitive connection between the drone and the operator's hand. In summary, we map the pose of the hand to the robot through the following equations:
% enabling the operator to participate in drone control 
% For position, we map the change in hand position to the aerial robot's reference according to the following equation:
\begin{equation}
\boldsymbol{p}_{B_c} = {\hat{\boldsymbol{p}}_B(t_0)} + \boldsymbol{k} \cdot \left({\hat{\boldsymbol{p}}_H(t)} -  {\hat{\boldsymbol{p}}_H(t_0)} \right), \quad {{^W_B\boldsymbol{q}}_c} = {^{W}_H\hat{\boldsymbol{q}}.}
\end{equation}
% where \( t_0 \) refers to the moment of entering this mode, \(\hat{\boldsymbol{p}}_B(t_0)\) and \(\hat{\boldsymbol{p}}_H(t_0)\) represent the initial positions of the drone and the operator's hand, respectively.
% % $\boldsymbol{k}$ $(\|\boldsymbol{k}\| \le 1)$ \revised{means scale factors for three axes}. 
% % % where a smaller value indicates a higher precision. 
% % We set \(\boldsymbol{k}\) to \([1, 1, 1]^T\), and note that it can be further adjusted according to the required accuracy of the task.
% For orientation, the hand attitude is directly mapped to the drone as
% \begin{equation}
% {{^W_B\boldsymbol{q}}_c} = {^{W}_H\hat{\boldsymbol{q}}.}
% \label{eq:rotation_update}
% \end{equation}
% % This equation updates the rotation quaternion \( {^{W}_B\boldsymbol{q}} \) of the drone by directly assigning the rotation quaternion \( {^{W}_H\hat{\boldsymbol{q}}} \) to \( {^{W}_B\boldsymbol{q}} \).

\begin{algorithm}[!t]
\caption{Our Shared Control Approach}
\label{alg:shard_control}
\begin{algorithmic}[1]

\Require 1) ${^W_H\hat{\boldsymbol{q}}}$, hand quaternion; 2) ${^W_B\hat{\boldsymbol{q}}}$, robot quaternion; 3) $^H\boldsymbol{\omega}$, hand angular velocity; 4) ${^W_H\boldsymbol{q}_{\text{start}}}$, initial hand quat to start the rotation; 5) $T_{\text{rot}}$, rotation period; 6) $\varepsilon_{\text{entry}}{=}0.2$, $\varepsilon_{\text{exit}}{=}0.1$, $\theta_{\text{trig}}{=}0.4$, $T_{\text{guard}}{=}5\,\mathrm{s}$, parameters for continuous rotation.
\Ensure ${^W_B\boldsymbol{q}}_c$ and $\boldsymbol{\omega}_c$, the target orientation and angular velocity of the aerial robot, respectively.
\While{\(\text{in } \textit{Operation Mode}\)}
    \State $\boldsymbol{q}_e \gets \textsc{QuatError}({^W_H\hat{\boldsymbol{q}}}, {^W_H\boldsymbol{q}_{\text{start}}})$;
    \State $\mathbf{v}_e \gets \mathrm{sign}(q_{e,w})\,[q_{e,x},\,q_{e,y},\,q_{e,z}]^T$;
    \If{\textbf{not} \textsc{IsRotating}()}
        \If{$\textsc{Align}(\mathbf{v}_{e,xy}, \mathbf{0}, \varepsilon_{\text{entry}})$}
            \If{$|\mathbf{v}_{e,z}| > \theta_{\text{trig}}$}
                \State $t_0 \gets \textsc{Now}()$;
                \State $\psi_0 \gets \textsc{ConvertToYaw}(\mathbf{v}_{e,z})$;
                \State $d_r \gets \mathrm{sign}(\mathbf{v}_{e,z}) \in \{+1,-1\}$;
            \EndIf
        \EndIf
        \State \Return $\{{^W_B\boldsymbol{q}}_c \gets {^W_H\hat{\boldsymbol{q}}}, \boldsymbol{\omega}_c \gets {^H\boldsymbol{\omega}}\}$;
    \EndIf
    \State $\tau \gets \textsc{Now}() - t_0$;
    \State $\boldsymbol{q}_{e,r2h} \gets \textsc{QuatError}({^W_B\hat{\boldsymbol{q}}}, {^W_H\hat{\boldsymbol{q}}})$;
    \State $\mathbf{v}_{e,r2h} \gets \mathrm{sign}(q_{e,r2h,w})\,[q_{e,r2h,x},\,q_{e,r2h,y},\,q_{e,r2h,z}]^T$;
    \State $c_1 \gets \textsc{Align}(\mathbf{v}_{e,r2h}, \mathbf{0}, \varepsilon_{\text{exit}})$;
    \State $c_2 \gets (\tau > T)$;
    \If{$(c_1 \lor c_2)\ \land\ (\tau > T_{\text{guard}})$}
        \State \textsc{Reset}();
        \State \Return $\{{^W_B\boldsymbol{q}}_c \gets {^W_H\hat{\boldsymbol{q}}}, \boldsymbol{\omega}_c \gets {^H\boldsymbol{\omega}}\}$;
    \EndIf
    \State $\omega \gets  2\pi / T \cdot d_r$;
    \State ${^W_B\boldsymbol{q}}_c \gets \textsc{EulerToQuat}(0,\ \pi/2,\ \omega\tau + \psi_0)$;
    % \State ${^W_B\boldsymbol{q}}_c \gets \textsc{HandleFlip}({^W_B\boldsymbol{q}}_c, {^W_B\hat{\boldsymbol{q}}})$
    \State $\boldsymbol{\omega}_c \gets (0,0,\omega)$;
    \State \Return $\{{^W_B\boldsymbol{q}}_c, \boldsymbol{\omega}_c\}$
\EndWhile
\end{algorithmic}
\end{algorithm}

The algorithm to trigger the continuous vertical rotation is listed in Alg. \ref{alg:shard_control}. In fact, shifting control authority to the robot is not challenging. In contrast, the most challenging part is determining \textit{when} the control authority should be returned to the operator.
Teleoperation systems should be resilient to users’ incorrect actions, including tremors and involuntary reflexes. However, when something goes wrong during manipulation, the operator should always know how to move the robot out of danger. In our case, we first use quaternion error to ensure there is no singularity problem in any situation, and then maintain the operator’s control over position throughout the process. We also ensure that the robot stops rotating if the orientation of the robot and hand become too close. We found that the robot’s behavior after autonomous turning is critical: if the robot suddenly jumps to the pose mapped from the hand, the operator may be startled. Hence, we try to set the stop orientation as close as possible to the hand.

\section{Experiments}  \label{sec:experiments}

\begin{figure}[t]
    \centerline{\includegraphics[trim=0 0 0 0,clip,width=3.3in]{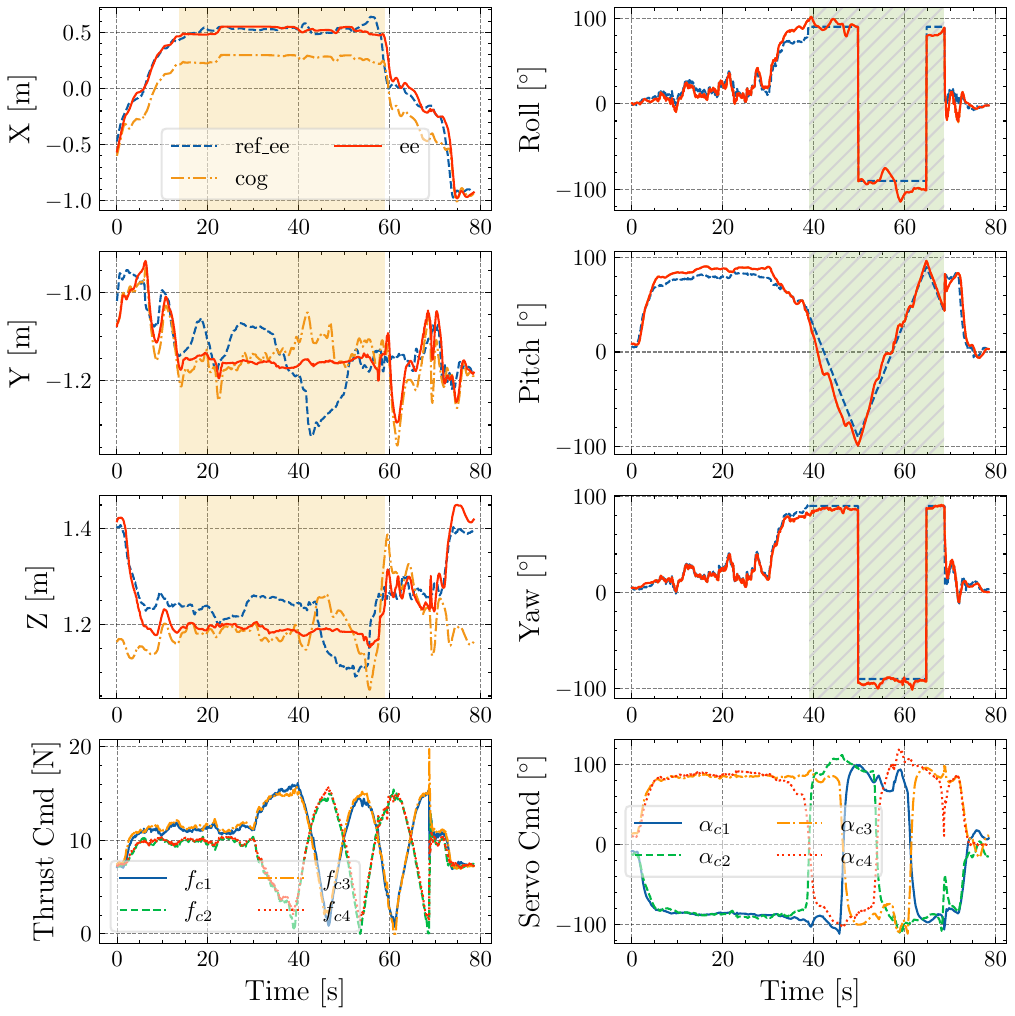}} 
    % 这个trim可以裁掉图片的白边
    % trim=left bottom right top
    \vspace*{-4mm}
    \caption{\textbf{Data of valve turning.} When the rotation of the human hand exceeds a threshold, the robot begins continuous rotation, with the range plotted as a green forward-slash area. The yellow area highlights the period when the robot is attached to the valve. Wrench estimation was shut down for safety. The RMSEs are: $p_x$: 0.0690\;m, $p_y$: 0.0622\;m, $p_z$: 0.0468\;m; roll: 7.8917$^\circ$, pitch: 9.0527$^\circ$, yaw: 4.1610$^\circ$.}
    \label{fig:exp_vertical_hand_valve_turn}
\end{figure}

We used a fork as the end-effector for valve turning, and an operator used the hand to control the robot. Since continuous valve rotation exceeds the natural flexibility of the human wrist and forearm, we introduced a \textit{shared-control} approach: when the operator’s hand rotation went beyond a threshold, the robot performed continuous vertical rotation for $360^\circ$. Although the attitude was governed, the position was always controlled by the operator for safety. As our framework did not include impedance control, the wrench compensation was also shut down to avoid excessive internal forces during turning.
The experimental data are shown in Fig. \ref{fig:exp_vertical_hand_valve_turn}.

From Fig. \ref{fig:exp_vertical_hand_valve_turn}, we can see that the robot can continuously rotate the vertically mounted valve. We observed an offset in position when the valve was engaged, indicating possible internal forces. Since the valve is rigid, internal forces accumulated through imperfect alignment may damage the end-effector, highlighting the need for compliance in the future. Additionally, the effector slipped out at the end of the yellow range, possibly due to uneven rotation caused by model error. More accurate calibration of the entire model may help ensure proper engagement.

% % \setlength{\dbltextfloatsep}{8pt plus 1.0pt minus 2.0pt}
% \begin{figure}[t] %bthp
%     \centering
%     % \vspace{-5mm}
%     \subfloat[Snapshots of turning a vertically installed valve.]{
%     % \hspace{-0.2cm}
%         \includegraphics[trim=3mm 3mm 0 0,clip,width=3.3in]{figs/exp_hand_valve_turn.jpg} % trim=left bottom right top
%         \label{fig:exp_hand_valve_turn_snapshot}
%     } \\
%     \subfloat[Data of turning a vertically installed valve.]{
%     % \hspace{-0.2cm}
%         \includegraphics[trim=0 0 0 0,clip,width=3.3in]{figs/exp_hand_valve_turn_data.pdf}
%         \label{fig:exp_hand_valve_turn_data}
%     }\vspace{-1mm}
%     \caption{An operator tries to turn the valve via a shared-control teleoperation system. When the rotating of human hand is beyond a threshold, the robot starts to continuously rotating, where the range are plotted as green forward-slash area. The yellow area highlights the period when the robot is attached to the valve. The wrench estimation is shutdown for safety. The RMSEs are: $p_x$: 0.0690\;m, $p_y$: 0.0622\;m, $p_z$: 0.0468\;m; roll: 7.8917$^\circ$, pitch: 9.0527$^\circ$, yaw: 4.1610$^\circ$.}
%     \label{fig:exp_vertical_hand_valve_turn}
% \end{figure}

\setlength{\textfloatsep}{8pt plus 1.0pt minus 2.0pt}

% % \setlength{\dbltextfloatsep}{8pt plus 1.0pt minus 2.0pt}
% \begin{figure}[t] %bthp
%     \centering
%     \subfloat[Poses of the operator's hand and the aerial robot in \textit{Operation Mode} during valve operation.]{
%         \includegraphics[trim=0 0 0 0,clip,width=3.5in]{figs/hand_turn_valve_v5.pdf}
%         \label{fig:mapping_mode_traj}
%     }\\
%     \subfloat[Absolute tracking error.]{
%         \includegraphics[trim=0 0 0 0,clip,width=3.3in]{figs/hand_turn_valve_abs_err.pdf}
%         \label{fig:mapping_mode_traj_err}
%     }
%     \caption{\textbf{Results for the accurate valve-turning task.} The yellow region highlights the valve-turning process, and the teleoperation latency is approximately 0.3\textasciitilde0.5\si{s}. The error increases right after the yellow area when the end effector is pulled away from the valve. \revised{The RMSEs are $p_x$: 0.0930\;\si{m}, $p_y$: 0.0519\;\si{m}, $p_z$: 0.0321\;\si{m}, $q_w$: 0.0944, $q_x$: 0.0467, $q_y$: 0.1122, $q_z$: 0.0816.}}
%     \vspace{-0.1cm}
%     \label{fig:accurate_exp_result}
% \end{figure}

\section{Conclusion}  \label{sec:conclusion}

In this work, we introduced the shared-control concept into our previous system, achieving continuous valve rotation in the real world. In the future, this philosophy could be extended to a general framework for human intent recognition and human–robot collaboration.

\clearpage % To start a new page
\bibliographystyle{bibtex/IEEEtran}
\bibliography{bibtex/IEEEabrv, bibtex/my_config, bibtex/references_pure}

%%%%%%%%%%%%%%%%%%%%%%%%%%%%%%%%%%%%%%%%%%%%%%%%%%%%%%%%%%%%%%%%%%%%%%%%%%%%%%%%

\end{document}